\documentclass{sig-alternate-05-2015}

\setcopyright{acmcopyright}
\usepackage{hyperref}
\usepackage{url}
\usepackage{color}
\usepackage{multirow} 
\usepackage{booktabs} 
\usepackage{array, graphicx}
\usepackage{color, colortbl}
\usepackage{pifont}
\usepackage[flushleft]{threeparttable}
\usepackage[font=small]{floatrow} 
\usepackage{subfig}
\usepackage{epstopdf}

\usepackage{etoolbox,siunitx} 
\robustify\bfseries
\newcommand{\boldentry}[2]{%
  \multicolumn{1}{S[table-format=#1,
                    mode=text,
                    text-rm=\fontseries{b}\selectfont
                   ]}{#2}}

\usepackage{caption} 
\newcommand*\rot{\rotatebox{90}}
\newcolumntype{L}[1]{>{\raggedright\let\newline\\\arraybackslash\hspace{0pt}}m{#1}}
\newcolumntype{R}[1]{>{\raggedleft\let\newline\\\arraybackslash\hspace{0pt}}m{#1}}

\definecolor{Gray}{gray}{0.9}
\newcommand*\OK{\ding{51}}

\begin{document}

\title{Comparative Study of Deep Learning Software Frameworks}

\numberofauthors{4} 
%
%
%
\author{
Soheil Bahrampour, Naveen Ramakrishnan, Lukas Schott, Mohak Shah\\
\and
 \affaddr{Research and Technology Center, Robert Bosch LLC}\\
 \email{$\lbrace \textrm{Soheil.Bahrampour, Naveen.Ramakrishnan, fixed-term.Lukas.Schott, Mohak.Shah}\rbrace @us.bosch.com$}
}

\maketitle

\begin{abstract}
Deep learning methods have resulted in significant performance improvements in several application domains and as such several software frameworks have been developed to facilitate their implementation. This paper presents a comparative study of five deep learning frameworks, namely Caffe, Neon, TensorFlow, Theano, and Torch, on three aspects: extensibility, hardware utilization, and speed. The study is performed on several types of deep learning architectures and we evaluate the performance of the above frameworks when employed on a single machine for both (multi-threaded) CPU and GPU (Nvidia Titan X) settings. The speed performance metrics used here include the gradient computation time, which is important during the training phase of deep networks, and the forward time, which is important from the deployment perspective of trained networks. For convolutional networks, we also report how each of these frameworks support various convolutional algorithms and their corresponding performance. From our experiments, we observe that Theano and Torch are the most easily extensible frameworks. We observe that Torch is best suited for any deep architecture on CPU, followed by Theano. It also achieves the best performance on the GPU for large convolutional and fully connected networks, followed closely by Neon. Theano achieves the best performance on GPU for training and deployment of LSTM networks. Caffe is the easiest for evaluating the performance of standard deep architectures. Finally, TensorFlow is a very flexible framework, similar to Theano, but its performance is currently not competitive compared to the other studied frameworks.
\end{abstract}

\section{Introduction}\label{sec:intro}

Deep learning methods have recently influenced several application domains, namely computer vision ~\cite{krizhevsky12, russakovsky14}, speech recognition~\cite{ding14, hannun14}, and nature language processing~\cite{collobert11}, where they have enjoyed significant performance improvements compared to state-of-art methods in the respective domains. For the latest list of domains and challenges on benchmark datasets where deep learning performed better than the existing state-of-art, see~\url{http://deeplearning4j.org/accuracy.html}. Most of the successful deep learning architectures are composed of a combination of different types of layers such as fully connected, convolutional, and recurrent layers and are usually trained with a variant of the stochastic gradient descent algorithm along with various regularization techniques such as dropout and weight decay~\cite{bengio15}.
As the popularity of the deep learning methods have increased over the last few years, several deep learning software frameworks have appeared to enable efficient development and implementation of these methods. The list of available frameworks includes, but is not limited to, Caffe, DeepLearning4J, deepmat, Eblearn, Neon, PyLearn, TensorFlow, Theano, Torch, etc. Different frameworks try to optimize different aspects of training or deployment of a deep learning algorithm. For instance, Caffe emphasises ease of use where standard layers can be easily configured without hard-coding while Theano provides automatic differentiation capabilities which facilitates flexibility to modify architecture for research and development. Several of these frameworks have received wide attention from the research community and are well-developed allowing efficient training of deep networks with billions of parameters, thanks to their strong GPU backends. Developers have constantly improved these frameworks by adding more features (e.g. by adding support for different types of convolution algorithms) and speed improvements to attract more users and foster research~\cite{bergstra11, collobert11torch7, vasilache14,  bastien12, jia14caffe}. Recently, the efficacy of several deep learning frameworks have been evaluated in~\cite{soumith15}. However, the comparison is only focused on speed performance of the convolutional frameworks. 
Hence, this paper expands on the previous benchmarks and evaluates five deep learning frameworks, namely: Caffe, Neon, TensorFlow, Theano, and Torch. Among the available software frameworks, Caffe, Theano, and Torch are indeed the top three well developed and widely used frameworks by the deep learning community. The reason for including Neon in this study is its recently reported state-of-the-art performance for training several deep learning architectures~\cite{soumith15}. Finally, TensorFlow has received much attention since its first release recently and we have included it in our benchmarking for completeness even though it does not yet~\footnote{as of the time of submission of this paper} officially support cuDNNv3, which is the latest official version supported by all the other studied frameworks. We evaluate these frameworks from the perspective of practitioners, on the following aspects:
\begin{itemize}
\item \textit{Extensibility}: Their capability to incorporate different types of deep learning architectures (convolutional, fully-connected, and recurrent networks), different training procedures (unsupervised layer-wise pre-training and supervised learning), and different convolutional algorithms (e.g. FFT-based algorithm).
\item \textit{Hardware utilization}: Their efficacy to incorporate hardware resources in either (multi-threaded) CPU or GPU setting.
\item \textit{Speed}: Their speed performance from both training and deployment perspectives.
\end{itemize}

The study will provide the users and enterprises a broad picture of the strengths and (current) limitations of the studied deep learning frameworks to enable them to assess suitability in the context of their requirements. Moreover, the discussions highlight the current limitations of the respective frameworks which can be addressed in their future developments~\footnote{Note that most of these frameworks have very active community support that keeps adding new features/functionalities potentially making some of our observations obsolete in the near future.}. We plan to share the code for all the frameworks in the near future through a publicly available webpage.

The rest of the paper is organized as follows: Section~\ref{sec:frameworks} gives a brief overview of the software frameworks we focus on in this paper; Section~\ref{sec:setups} describes the benchmarking set up which is followed by results and conclusions in Section~\ref{sec:results} and Section~\ref{sec:conclusions}, respectively.

\section{Overview of the deep learning frameworks}\label{sec:frameworks}

\begin{table*}[!t]
\caption{Community involvements for some of the deep learning frameworks as of 02/08/2016.}
\label{tab:FrameworksStats}
\centering
\small
\setlength{\tabcolsep}{1.5mm}
\begin{tabular}{lrrrrrrr}
 Measures & Caffe & DeepLearning4J & Eblearn & Neon & TensorFlow & Theano & Torch7  \\
\toprule
$\textrm{Number of members in Google groups}$ & 4220 & 857 & 109 & 73 & 661 & 2827 &1874 \\
$\textrm{Number of contributors in GitHub}$ & 172 & 57 &  NA &  31 & 81 & 207 & 77\\
\bottomrule
\end{tabular}
\end{table*} 

\begin{table*}[!t]
\caption{Properties of Caffe, Neon, TensorFlow, Theano, and Torch as of 02/08/2016.}
\label{tab:FrameworksComparison}
\centering
\small
\begin{tabular}{L{0.9in}L{0.6in}L{1in}L{1in}L{1in}L{0.6in}L{0.01in}}
 Property & Caffe & Neon & TensorFlow & Theano & Torch & \\
\toprule
\rowcolor{Gray} Core & C\texttt{++}  & Python & C\texttt{++} & Python  &  Lua & \\[2ex]
CPU & \OK & \OK &  \OK & \OK  & \OK &\\ [2ex]
\rowcolor{Gray} Multi-threaded CPU & \OK Blas & \textbf{x} Only data loader & \OK Eigen & \OK Blas, conv2D, limited OpenMP & \OK \textbf{Widely used} &\\[1ex]
GPU & \OK & \OK customized Nvidia backend &  \OK &\OK & \OK  &\\[2ex]
\rowcolor{Gray} Multi-GPU & \OK (only data parallel) & \OK  & \OK \textbf{Most flexible} &  \textbf{x} Experimental version available & \OK &\\[2ex]
 Nvidia cuDNN & \OK & \textbf{x} & \OK & \OK  & \OK & \\[2ex]
\rowcolor{Gray} Quick deploy. on standard models & \OK \textbf{Easiest} & \OK & \OK & \textbf{x} Via secondary libraries  & \OK  & \\[1ex]
 Auto. gradient computation & \OK & \OK Supports Op-Tree & \OK & \OK \textbf{Most flexible} (also over loops)  &  \OK & \\[1ex]

\bottomrule
\end{tabular}
\end{table*} 

With deep learning methods gaining popularity in many applications domains over the last few years, there have been quite a lot of interest from many academic (e.g. Univ. of California Berkeley, NYU) and industry groups (e.g. Google, Facebook) to develop software frameworks (e.g. Theano, Caffe) that help easily create and test various deep architectures. At the time this paper was written, some of the widely used software frameworks for deep learning were: Caffe, Theano, Torch, Neon, TensorFlow, Chainer, DeepLearning4J, deepmat, Eblearn, MXNet, etc. (for a more complete list of Deep Learning Softwares see \url{http://deeplearning.net/software_links/}). Many of these frameworks are mature already as of today and are very fast in training deep networks with billions of parameters – thanks to their strong CUDA backends. Today, almost every group training deep networks use Graphical Processing Units (GPU) to accelerate the training process and this has led to joint development of software libraries (e.g. cuDNN) between academic (e.g. Berkeley, NYU) and industry players (e.g. Nvidia). Table~\ref{tab:FrameworksStats} shows the number of users in Google groups and the number of contributors\footnote{We only report the number of the contributors in the main repository of the framework. The numbers do not include any other relevant repositories.} for each of the frameworks in their corresponding GitHub repositories. It is clear that the top three widely developed and supported deep learning frameworks are Caffe, Theano, Torch, and are thus selected in this paper for the benchmarking purposes. We also evaluated Neon framework from Nervana as it has recently shown the state-of-the-art performance for training convolutional networks~\cite{soumith15}. We have also included TensorFlow from Google in our experiments as it has recently received much attention.
 Table~\ref{tab:FrameworksComparison} shows the general properties of these five deep learning frameworks. Note that, for this paper, we restrict ourselves to frameworks suitable for single node (with potentially multiple GPUs) but not distributed deep learning frameworks like DeepLearning4J. For a brief review of the selected frameworks see Appendix.

\section{Benchmarking setup}\label{sec:setups}

\subsection{Evaluation Metrics}\label{ssec:metrics}
We use the two following evaluation metrics to obtain a holistic understanding of speed of the five deep learning frameworks under various system scenarios and application domains:
\begin{itemize}
\item Forward Time: We measure the time it takes for an input batch of a pre-selected batch size, for a given dataset and network, to flow through the entire network and produce the corresponding output. This is important as it indicates the latency of a deep network when deployed in real-world.
\item Gradient Computation Time: We also measure the time it takes to get the gradients for each measurable parameter in the deep network for a given input batch. This is an important indicator of the training time. Note that, for most of the cases (e.g. Torch), this gradient computation time is the summation of the times spent in calling the corresponding \textit{forward} and \textit{backward} functions as these two functions should be called consecutively to compute the gradients. But for Theano, this gradient computation time is measured by calling a Theano function that is compiled to generate the gradients given the inputs to the networks which \textit{implicitly} performs the forward and backward steps through computational graphs. It should be noted that the gradient computation time we report, does not include the time taken to update the network parameters, such as computation of learning rate, weight decay, momentum term, etc.
\end{itemize}
For Theano, one initially need to compile forward and gradient computation functions before calling them during execution. To provide a complete picture, these compilations times are also reported (See Tabel~\ref{tab:theanoCompilation}). We also report the GPU memory usage for large networks.

\subsection{System setup}\label{ssec:system}
All the experiments are performed on a single machine running on Ubuntu 14.04 with Intel Xeon CPU E5-1650 v2 @ 3.50GHz × 12; Nvidia GeForce GTX Titan X/PCIe/SSE2; 32 GiB DDR3 memory; and SSD hard drive. We used openCV 3.0. and OpenBLAS 0.2.14 with commit ID 3684706. For Caffe, commit ID 8c8e832 is used. For Neon, version 1.0.0.rc1 (2015-09-08) with the commit ID of a6766ff is used. For TensorFlow, we installed version 0.6.0 using pip installation. Theano version 0.7.0.dev and Torch7 used here have commit IDs 662ea98 and 8c8e832, respectively. The commit ID for fbcunn is 5bb9785. For Caffe, Theano and Torch, we used CUDA 7.5 and cuDNN v3 while for TensorFlow we used CUDA 7.0 and cuDNN v2 since these are the officially supported libraries at the time of submission. Data arrays are stored using the float32 format.

\section{Results and discussions}\label{sec:results}
The evaluations are performed by training stacked autoencoders and convolutional networks on the MNIST~\cite{lecun98} and the ImageNet datasets~\cite{deng09} as well as training LSTM network using the IMDB review dataset~\cite{maas11}. Note that the evaluation metrics can vary drastically based on the CUDA package used along with the native software. For example, in Torch, one can perform the convolution operations using Nvidia cuDNN library or cunn library (a CUDA backend for the nn package) or fbcunn library (deep learning CUDA extensions from Facebook AI Research containing FFT based fast convolutions). In Theano, it is also straightforward to perform convolution using cuDNN or conv-fft. The conv-fft is a FFT-based implementation of convolution operation on Theano. Hence we try to use as many libraries as possible for each of the cases and measure the performance to present the inherent tradeoffs with each of the libraries. We use the same blas library for Caffe, Theano, and Torch which performs majority of the computations when CPU is used. Neon uses its own CPU/GPU backend. Moreover, wherever applicable, we measure the speeds with both GPU and CPU (single and multi-threaded) so as to understand the hardware specific behaviours of these frameworks from both training and deployment perspectives. We perform several iterations of warm-ups before timing the operations. The timings reported here are average of 20-1000 iterations and are controlled to have small standard deviations. 

\subsection{LeNet}\label{ssec:lenet}

\begin{figure*}
\CenterFloatBoxes
\begin{floatrow}
\ttabbox
{
\begin{tabular}{l@{\hspace{0.8\tabcolsep}}l@{\hspace{0.8\tabcolsep}}l@{\hspace{0.8\tabcolsep}}r@{\hspace{0.8\tabcolsep}}r}
\multicolumn{3}{c}{Setting}  & Gradient (ms) &  Forward (ms) \\
 \cmidrule(lr){1-3}  \cmidrule(lr){4-4}  \cmidrule(lr){5-5}
 \multirow{11}{*}{\rot{CPU}} & \multirow{4}{*}{\rot{1}} & Neon & 545.6 & 172.7\\
&& TensorFlow & 93.8 & 42.1   \\ 
&& Theano & 141.1&48.3   \\
&& Torch & 46.1&18.1  \\
 \cmidrule(lr){2-5}
&\multirow{3}{*}{\rot{6}} & TensorFlow & 45.8& 16.8\\
& & Theano & 142.7 & 50.4 \\
& & Torch  & 18.1 & 5.6  \\
 \cmidrule(lr){2-5}
&\multirow{4}{*}{\rot{12}} & Caffe&66.4& 33.7 \\
&& TensorFlow & 50.1 &16.4   \\
&& Theano & 204.3 &78.7   \\
&& Torch & \textbf{16.5}&\textbf{4.6}  \\
\cmidrule(lr){1-5}
\multirow{8}{*}{\rot{GPU}} & \multicolumn{2}{l}{Caffe + cuDNN v3} & 1.9 & 0.8 \\
& \multicolumn{2}{l}{Neon} & 2.3 & 1.0 \\
& \multicolumn{2}{l}{TensorFlow + cuDNN v2} &  14.6& 4.5  \\
& \multicolumn{2}{l}{Theano + cuDNN v3} & \textbf{1.4} & \textbf{0.5} \\
& \multicolumn{2}{l}{Theano + conv-fft} & 5.6 & 2.7\\
& \multicolumn{2}{l}{Torch + cuDNN v3} & 1.7 & \textbf{0.5} \\
& \multicolumn{2}{l}{Torch + cunn} & 13.6 & 5.8 \\
& \multicolumn{2}{l}{Torch + fbcunn} & 2.1 & 0.9 \\
\bottomrule
\end{tabular}
}
{
\caption{The averaged processing times using batch size of 64.}
\label{tab:lenet}
}
\killfloatstyle

\ffigbox[6cm]
  {\caption{\small The averaged processing times for LeNet on GPU using different batch sizes. The cuDNN v2 is used for TensorFlow and cuDNN v3 is used for Caffe, Theano, and Torch.}\label{fig:lenetbatches}}
  {\includegraphics[scale=0.23]{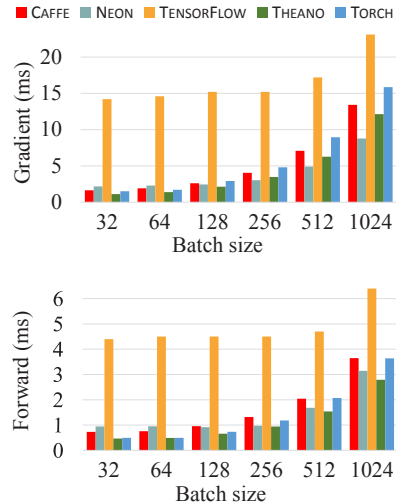}} 
\end{floatrow}
\end{figure*}

The first benchmark is a slightly modified LeNet neural network on the MNIST dataset~\cite{lecun98} where the sigmoid activations are replaced with \textit{ReLU} units and softmax logistic loss layer is used instead of the RBF network. It consists of two convolution-pooling layers with the \textit{tanh} activation functions and two fully connected layers. For Caffe, the network is available in Caffe model repository. For Theano, the code is adopted from~\cite{lisa14} while the TensorFlow code is adopted from~\cite{soumith15}. For Torch, we used the ``mnist'' package for easily loading the dataset and wrote our own script for timing purposes. For Neon\footnote{We directly call the \textit{fprop} and \textit{get\_cost} functions to time the forward pass. Similarly, the \textit{get\_errors} and \textit{bprop} functions are used to time the backward pass.}, we adopted the code from the Neon GitHub repository. Neon requires the kernel size of convolutional layers and mini-batch size to be multiples of 4 and 32, respectively when employed on GPU. Thus the second convolution layer for Neon implementation is chosen to have 52 filters instead of 50 filters used in the other frameworks. 

Table~\ref{tab:lenet} shows the averaged processing time for gradient computation as well as the time for a forward step obtained by the five frameworks on both CPU and GPU using batch size of 64. For CPU timings, the number of threads used in each experiment is also reported. It should be noted that Neon cannot be configured to use multiple CPU threads and thus its performance on CPU is only reported with one thread. On the other hand, Caffe can be configured only during installation to run on a pre-determined number of threads ($12$ here) and thus it's performance on CPU is only reported with $12$ threads. TensorFlow, Theano, and Torch are flexible in selecting the number of threads and thus their performances on CPU are reported with multiple settings. We report results for six and twelve threads since our system has six physical cores which can also be used with twelve threads using Hyper-Threading. When GPU is used, the underlying convolution library (e.g. cuDNN) is mentioned along with the framework. Neon uses its own GPU/CPU backend as mentioned before. The processing times clearly show the advantage of GPU over CPU for training deep convolutional networks. This advantage would be more significant when training more complex models with larger data as will be shown later in this section. Torch results in best performance when comparing CPU times while Neon results in the worst performance on CPU. It is seen in the GPU experiments that cuDNN is faster for this network compared to the conv-fft. In general, the performance gain of using the FFT-based approach is highly dependent on the size of the input and kernel sizes~\cite{mathieu13}. Theano results in best performance for the gradient computation on GPU while Torch and Theano achieve the best GPU performance for deployment. TensorFlow results in worst performance on GPU. One reason might be that it uses cuDNN v2 while Caffe, Theano, and Torch use cuDNN v3. It should also be noted that MNIST is a relatively small dataset and fits on the CPU host memory or the GPU device memory. Therefore, when Theano, Torch, or Neon is employed on GPU, the data is entirely copied into the GPU memory once before the training starts to avoid possible delays caused by communications between GPU and host for copying mini-batches\footnote{This is done on Theano using shared variables, on Torch by calling the \emph{:cuda()} function, and on Neon using the DataIterator class. On TensorFlow, this can be done by appropriately stetting trainable parameter when defining variables. Copying of the entire dataset into the memory can also be done for Caffe using MemoryData layer. Here we used efficient LMDB database for Caffe and the communication overhead is not significant. The combined averaged forward and backward computational time of the data layer of LeNet in Caffe is about 1/1220 (1/30) of the total computational time of the batch when using CPU (GPU). This includes the time to rescale the images of the batch to the unit range.}.  

Figure~\ref{fig:lenetbatches} shows the gradient computation time and forward step time of the five frameworks on GPU using different batch sizes. It is seen that Theano has the best gradient computation time for small batches while Neon has the best performance for large batches. Theano consistently has the minimum forward time, specially for the large batch sizes. It is seen that Torch and Caffe performances drop more rapidly as the batch size increases. It is also seen that TensorFlow results in the worst performance, specially for small batch sizes.

%

\subsection{AlexNet}\label{ssec:alexnet}

\begin{figure*}
\CenterFloatBoxes
\begin{floatrow}
\ttabbox
{
\begin{threeparttable}
\sisetup{detect-weight=true,detect-inline-weight=math,output-decimal-marker=\textnormal{.}}
\begin{tabular}{l@{\hspace{0.5\tabcolsep}}l@{\hspace{0.1\tabcolsep}}S[table-format=5.1]@{\hspace{0.1\tabcolsep}}S[table-format=5.1]@{\hspace{0.1\tabcolsep}}S[table-format=1.1]}
\multicolumn{2}{c}{\multirow{2}{*}{Settings}}  & {Gradient} &  {Forward} & {GPU RAM}\\
& & {(ms)} & {(ms)} & {(GB)} \\
 \toprule
 \multirow{5}{*}{\rot{CPU}} & Caffe (12 threads) & 43152 & 19817 & {-}\\
& Neon (1 thread)$^* {}^{\dagger}$ & 100987 & 28828 & {-}\\
& TensorFlow (12 threads)$^* {}^{\dagger}$ & 15560 & 4631 & {-}\\
& Torch (6 threads)$^*$ &11977 &4383 & {-}\\
& Torch (12 threads)$^*$ & \boldentry{4.0}{8421}  & \boldentry{6.1}{2746}& {-}\\
 \cmidrule(lr){1-5}
\multirow{12}{*}{\rot{GPU}} & Caffe + cuDNN v3 & 422.4 & 111.7 & 4.1\\
& Theano + cuDNN v3 & 529.8 & 162.8 & \textbf{3.3}\\
& Theano + cuconv & 684.9 & 156.1 & 5.6\\
& Torch + cuDNN v3 & \boldentry{5.1}{390.2} & \boldentry{5.1}{92.5}& 3.7\\
 \cmidrule(lr){2-5}
& Caffe + cuDNN v3$^* {}^{\dagger}$ & 521.2 & 130.4  & 2.7 \\ 
& Neon$^* {}^{\dagger}$ & 290.5 & \boldentry{5.1}{96.3} & \textbf{2.4}\\
& TensorFlow + cuDNN v2$^* {}^{\dagger}$ &  742.0&220.0  & 3.9 \\
& Theano + cuDNN v3$^* {}^{\dagger}$ & 561.2 & 172.3 & 2.7\\
& Theano + cuconv$^* {}^{\dagger}$ & 698.8 & 211.1  & 6.8 \\
& Torch + cuDNN v3$^* {}^{\dagger}$ & 405.9 & 100.7 & 2.8\\
& Torch + cunn$^* {}^{\dagger}$ & 915.7 & 365.3 &2.9\\
& Torch + fbcunn$^* {}^{\dagger}$ & \boldentry{5.1}{286.3} & 98.4 & 4.8\\
\bottomrule
\end{tabular}
\begin{tablenotes}
      \small
      \item $^*$Without local response normalization layers.  
       \item $^{\dagger}$ No grouping is performed in convolutional layers. 
    \end{tablenotes}
 \end{threeparttable}
} 
{
\caption{The averaged processing times for AlexNet as well as peak GPU memory usage with batch size of 256.}
\label{tab:imagenet}
}
\killfloatstyle

\ffigbox[5.1cm]
  {\caption{\small The averaged processing times for AlexNet on GPU using different batch sizes. The cuDNN v2 is used for TensorFlow and cuDNN v3 is used for Caffe, Theano, and Torch. Experiments are without normalization layers and grouping.}\label{fig:alexnetbatches}}
  {\includegraphics[scale=0.23]{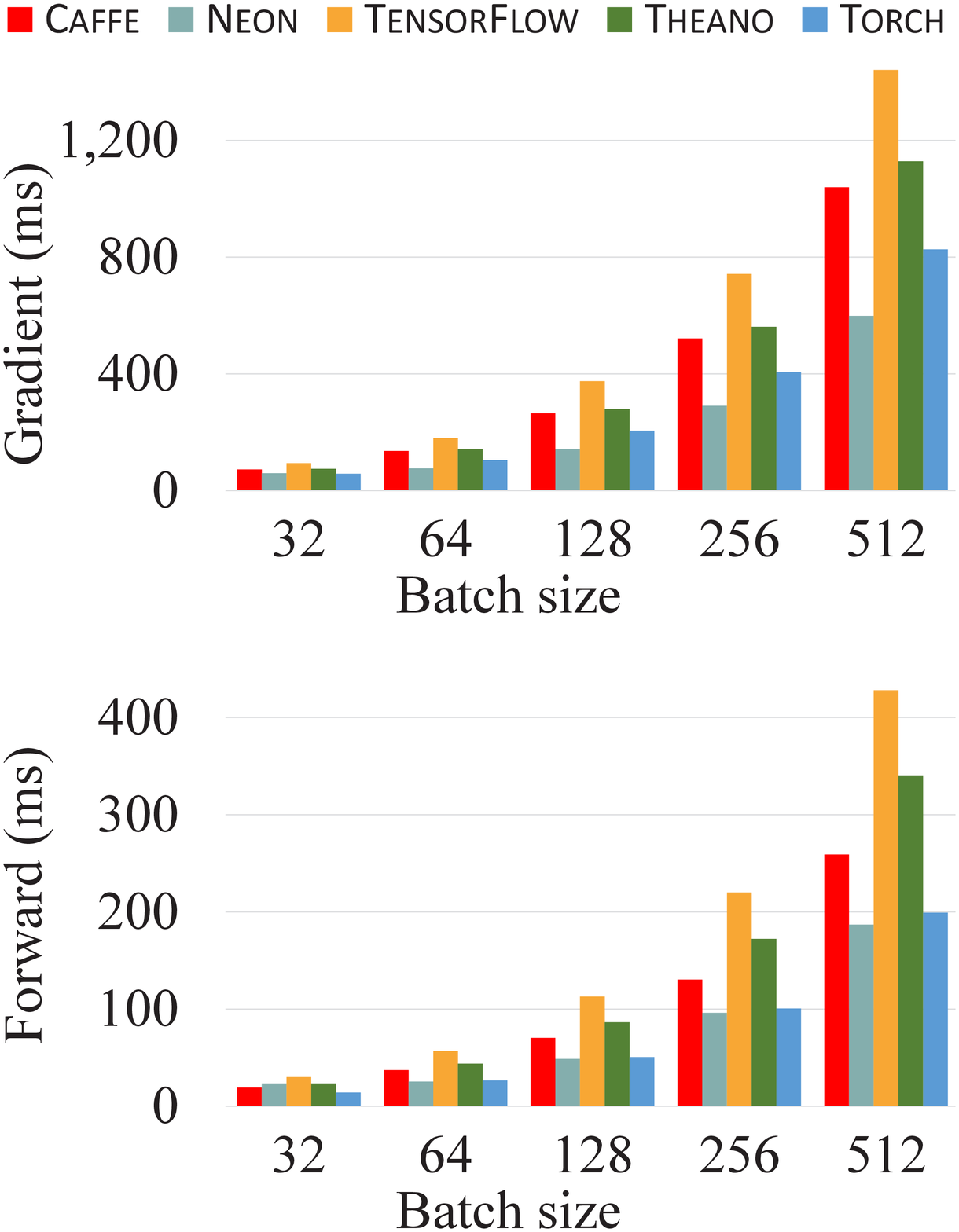}} 
\end{floatrow}
\end{figure*}

In this section, we train AlexNet~\cite{krizhevsky12} on the ImageNet dataset. Note that there have been many recent, larger networks like GoogleNet, OxfordNet, etc. but we stick with AlexNet as it is the first network that significantly improved performance on ImageNet and is very popular. The network consists of five convolution layers, out of which three of them use grouping which restrict the connectivity of filters, and two have local response normalization (LRN) layers. The networks also has three pooling layers, two fully connected layers with ReLU activation units and dropout, and a softmax logistic loss. Each image is cropped to have dimension of 224. The data augmentation using random cropping or transformation is not performed\footnote{The ImageNet data is accessed in Caffe using the LMDB database, in Neon using $ImgMaster$ class, in Theano using Hickle, and in Torch using a multithreaded data loader provided in~\cite{soumith15_multigpu} that creates a pre-specified number of threads for parallel data loading from disk.}. For Caffe and Neon, the network is available from the corresponding GitHub repository. Neon currently does not support grouping and LRN layers. For TensorFlow, the code is adopted from~\cite{soumith15}. TensorFlow does not currently support grouping. For Theano, the code of~\cite{ding14} is adopted without performing data parallelization. We updated the implementation to avoid unnecessary \textit{dimshuffle} operations. The convolution on GPU on Theano is performed by calling either the $dnn.dnn\_conv$ function from cuDNN library or the corresponding function from pylearn2 cuda-convent wrapper\footnote{The cuda-convnet~\cite{krizhevsky12} is a fast implementation of convolution but has some restrictions on input and kernel shapes with a different memory layout compared to Theano convolution operator.}\footnote{The Theano implementation does not use the standard convolution function ($conv.conv2d$) as it does not implement the type of padding used in AlexNet, known as ``same" padding. Thus, it is not possible to perform the AlexNet Theano experiment on CPU or using conv-fft as they can be accessed through the conv2d function.}. The latter is referred as cuconv in the results. 
 For Torch, in addition to cuDNN library, we report the timings on GPU using both cunn and fbcunn libraries. Note that fbcunn does not support stride lengths greater than 1. So when reporting fbcunn results, we use the cuDNN-based convolution for the first layer of AlexNet and fbcunn-based convolutions for the rest. Furthermore, cunn and fbcunn do not support grouping. In addition to reporting the results for the exact AlexNet implementation, we also report the results without LRN layers and with grouping set to one to make the comparison transparent.

Table~\ref{tab:imagenet} shows the performance of the five frameworks on the AlexNet using batch size of 256.
 To have a better performance comparison across the frameworks, the time required for data loading and processing (mean normalization) in each batch is excluded from time of forward and backward steps in all our experiments. We also report the peak GPU memory consumption to illustrate the efficacy of the frameworks in implementing deep networks\footnote{We used \textit{nvidia-smi} to monitor the GPU memory consumption.}. In the CPU setting, Torch results in the best speed performance, similar to the LeNet results. The speed-up obtained by using GPU instead is more significant (at least $25\times$) here compared to the LeNet. When employed on GPU, Torch results in the best performance on the exact AlexNet implementation. When LRN layers are dropped and grouping are set to one, Torch using fbcunn results in the best gradient computation performance while Neon results in the best forward pass performance followed closely by Torch. Similar to LeNet experiment, TensorFlow results in worst performance when employed on GPU. Figure~\ref{fig:alexnetbatches} shows the performance of the five frameworks on GPU using different batch sizes when no LRN layers are used here and grouping for all convolutional layers are set to one. Note that Neon and Torch have consistent superior performances for different batch sizes for the forward pass but Neon results in best performance for the gradient computation time. In terms of GPU memory consumption, similar efficient usage are observed across Caffe, Theano, and Torch when cuDNN is used while Neon has the most memory efficient usage. TensorFlow has the highest memory consumption. We also noticed from our experiments that the LMDB database used in Caffe has significantly better performance than the other data access layers used in Neon, Theano and Torch frameworks~\footnote{Data access layer is not tested on TensorFlow.} as it supports concurrent reads. Caffe also uses pre-fetching to eliminates IO latency. This can be an area of future developments for Neon, Theano, and Torch to make the LMDB database (or other efficient databases) and pre-fetching available in their frameworks\footnote{Pre-fetching and multi-thread processing can also be implemented in Torch as has been done for the Imagenet example in~\cite{soumith15_multigpu}.}.

\subsection{Stacked autoencoders}\label{ssec:sae}

\begin{table*}[!t]
\caption{The averaged processing times of the stacked autoencoders (AE) for both pre-training and fine-tuning steps using batch size of 64. The encoder dimensions for AE1, AE2, and AE3 are 400, 200, and 100, respectively. For the unsupervised pre-training step, the gradient computation times are reported for the individual AEs along with the total gradient computation. For the supervised fine-tuning step of the stacked enocoders (SE), both gradient computation and forward pass times are reported. Caffe and Neon implementations do not have tied weights.}
\label{tab:sae1}
\centering
\small
\setlength{\tabcolsep}{2mm}
\sisetup{detect-weight=true,detect-inline-weight=math,output-decimal-marker=\textnormal{.}}
\begin{tabular}{lllS[table-format=2.1]S[table-format=2.1]S[table-format=2.1]S[table-format=2.1]S[table-format=2.1]S[table-format=2.1]S[table-format=2.1]}
& &  & \multicolumn{5}{c}{Gradient (ms)} &  {Forward (ms)}\\
 \cmidrule(lr){4-8} \cmidrule(lr){9-9}
\multicolumn{3}{c}{Setting} & {AE1} & {AE2} & {AE3} & {Total pre-training} & {SE} & {SE} \\
\toprule
\multirow{8}{*}{\rot{CPU threads}} & \multirow{3}{*}{\rot{1}} & Neon & 14.6 & 11.5 & 10.5 & 35.6 & 9.7 & 4.8 \\
&& TensorFlow & 32.4 & 35.9 & 47.1 & 115.4 & 70.2 & 51.9 \\
&& Theano & 14.8 & 10.5 & 8.4 & 33.7 & 8.2 & 6.4 \\
&& Torch & 13.7 & 8.7 & 6.5 & 28.9 & 8.2 & 5.0 \\
 \cmidrule(lr){2-9}
& \multirow{2}{*}{\rot{6}} & TensorFlow & 18.7 & 28.4 & 41.2 & 88.3 & 62.0 & 46.4 \\
&& Theano & 5.8 & 3.9 & 2.5 & 12.2 & \boldentry{2.1}{2.6} & \boldentry{2.1}{1.8} \\
&& Torch & \boldentry{2.1}{5.0} & \boldentry{2.1}{3.0} & \boldentry{2.1}{2.3} & \boldentry{2.1}{10.3} & 3.3 & 1.9 \\
 \cmidrule(lr){2-9}
&\multirow{3}{*}{\rot{12}} & Caffe&11.7&10.6&8.6 &30.9 &7.4 &6.1 \\
&& TensorFlow & 18.0 & 28.0 & 40.9 & 86.9 & 63.8 & 46.3 \\
&& Theano & 6.2& 4.4 & 4.0 & 14.6 & 3.7 & 2.8 \\
&& Torch & 9.8& 4.2& 3.2 & 17.2 & 3.8 & 2.3 \\
\cmidrule(lr){1-9}
 \multirow{4}{*}{\rot{GPU}} & \multicolumn{2}{l}{Caffe + cuDNN v3} & 0.7 & 0.8 & 0.8 & 2.3 & 1.0 & 0.6\\
& \multicolumn{2}{l}{Neon} & 1.1 & 1.5 & 1.7 &4.3 & 1.8 &0.9  \\
& \multicolumn{2}{l}{TensorFlow + cuDNN v2} & 11.9 & 23.9 & 38.1 & 73.9 & 57.9 & 44.7 \\
& \multicolumn{2}{l}{Theano + cuDNN v3} &\boldentry{2.1}{0.6} & \boldentry{2.1}{0.4}& \boldentry{2.1}{0.3}& \boldentry{2.1}{1.3}& \boldentry{2.1}{0.4}&\boldentry{2.1}{0.2} \\
& \multicolumn{2}{l}{Torch + cuDNN v3} & \boldentry{2.1}{0.6} & 0.5 & 0.5& 1.6&0.7 &0.3\\
\bottomrule
\end{tabular}
\end{table*}

To benchmark a scenario with layer-wise pre-training procedure,
 we choose stacked autoencoders. This also provides a better picture of the performances of different frameworks when only fully-connected layers are used. We train three autoencoders (AEs) where each AE has a encoder and a corresponding decoder layer with tied weights, i.e. the decoder weights are transpose of the encoder weights.
 The sigmoid activation functions are used. The network is trained on the MNIST dataset in two steps: layer-wise unsupervised training and supervised fine-tuning. The unsupervised layer-wise training step is performed similar to the procedure in~\cite{bengio07} using mean squared loss function. The AE1 is first trained on the raw images and then its weights are fixed. The AE2 is then trained on the resulting outputs of the first encoder and this procedure is repeated until all AEs are trained. Note that once an AE is trained, its encoder outputs are not computed and recorded for the entire dataset in the memory to be used for the following AE\footnote{Saving the outputs of trained encoder for the entire input would improve computational time but is not a memory-efficient procedure, specially for large datasets, and therefore is not employed here.}, rather each batch is separately processed. Thus the forward pass for AE2, for example, includes a pass from raw image data to the first encoder and AE2 before loss is computed. In the supervised fine-tuning step, the training is performed on the stacked encoders (SE) of each AE with a softmax layer of size 10 and a cross entropy loss function. The decoders are not present in this fine-tuning step. 

The above pre-training and fine-tuning steps are implemented in Theano~\cite{lisa14}, TensorFlow and Torch. For Caffe, pre-training step is implemented using a few tricks. We have four configuration files in which three of them handle training of the individual AEs and one handles the fine-tuning step on the SE. We set the learning rates of the layers that should not be updated during pre-training step to zero\footnote{One can use PyCaffe and the Caffe ``net surgery" procedure to transfer the learned weights of each trained AE to the following AE. This is not performed here as we are only interested in the computational performance.}. For example, when training the second AE, the learning rate for the weights of first encoder are set to zero\footnote{It should be noted that Caffe detects the zero learning rates and does not perform unnecessary calculations.}. Our Neon implementation is very similar to Caffe implementation and multiple optimizers are used to set the learning rates of the layers that should not get updated to zero. It should be noted that Caffe and Neon do not yet support tied weights and thus, different from our Theano, TensorFlow and Torch implementations, have independent parameters for encoders and decoders. The performance of the five frameworks are shown in Table~\ref{tab:sae1} where the encoders of the three AE layers have 400, 200 and 100 hidden layers, respectively. It is seen that Torch and Theano results in superior performance and TensorFlow followed by Neon results in the worst performance for both CPU and GPU settings. We have also evaluated the frameworks in a different setting, where the number of hidden layers of encoders of AE1, AE2 and AE3 are 800, 1000 and 2000, respectively. For this larger network, Caffe results in better performance compared to Theano on GPU but Torch again achieves the best performance. The results of this experiment are shown in Table~\ref{tab:sae2} in the Appendix.

\subsection{LSTM}\label{ssec:lstm}
In this section, we train a LSTM network~\cite{graves12} for the task of sentiment analysis on IMDB dataset. In this task, each sentence is considered as a (varying-length) sequence of words. The network architecture is the same as the one used in~\cite{lisa14}. It consists of an embedding layer followed by an LSTM layer. The outputs of the LSTM layer are then averaged and fed to a linear fully connected layer with softmax logistic regression for binary classification. The sequences within each batch are padded to have the same size as the largest sequence within the batch and a masking array is used to make sure the recursive computations of the LSTM layer remain valid. For Torch, we use the LSTM layer from the ``rnn'' package~\cite{rnn_torch} along with the \emph{MaskZero} and \emph{LookupTableMaskZero} modules for handling the varying length scenario.

Caffe does not yet officially support cyclic architectures, and in particular LSTM, and thus its performance is not reported here\footnote{Recently, a pull request is submitted to the official Caffe repository which adds the support for RNN and LSTM. See \url{http://jeffdonahue.com/lrcn/} for more information.}. While Neon and TensorFlow have LSTM layers\footnote{Neon has the option to pad data to fixed sizes. TensorFlow supports the idea of bucketing and padding variable length sentences and forming batches that contains equal-sized sequences}, they do not accept variable length inputs within a batch and thus are not used here. It should be noted that one of the main advantages of recurrent networks are their capabilities in handling variable length inputs without the need to make the window size constant~\cite{graves12}.

We used 124 iterations, one entire epoch, to average the computational time for different padding sizes. Also shuffling is not performed on the training set to make sure different frameworks receive the same sequence of batches and thus have the same number of flops. As the dataset is small, it is initially loaded into the device or host memory. Table~\ref{tab:lstm} shows the performance of Theano and Torch for the LSTM network.
 As with previous cases, Torch performs best for CPU but with a GPU, Theano results in better performance. 

\section{Conclusions}\label{sec:conclusions}
We evaluated five of the top deep learning frameworks, namely Caffe, Neon, TensorFlow, Theano and Torch for a variety of settings on a single machine. Here are our main observations:  
\begin{itemize}
\item Theano and Torch are the most extensible frameworks both in terms of supporting various deep architectures but also in terms of supported libraries. The symbolic differentiation is one of the most useful features that Theano offers for implementing non-standard deep architectures. Torch community is trying to fill this gap\footnote{For more information see~\url{https://blog.twitter.com/2015/autograd-for-torch}}.
\item For CPU-based training and deployment of \emph{any} tested deep network architecture, Torch performs the best followed by Theano, and Neon has the worst performance.
\item For GPU-based deployment of trained convolutional and fully connected networks, Torch is best suited, followed by Theano.
\item For GPU-based training of convolutional and fully connected networks, we noticed Theano is fastest for small networks and Torch is fastest for larger networks. Neon is very competitive on GPU for large convolutional networks.
\item For GPU-based training and deployment of recurrent networks (LSTM), Theano results in the best performance.
\item Torch could greatly benefit from expanded documentation of its libraries and capabilities and better error debugging tools.
\item TensorFlow is a very flexible framework, specially in employing homogeneous/heterogeneous devices for the various parts of the computational graph. However, its performance on a single GPU is not as competitive compared to the other studied frameworks.
\end{itemize}

\begin{table}[!t]
\caption{The averaged processing times of the LSTM using batch size of 16.}
\label{tab:lstm}
\centering
\small
\setlength{\tabcolsep}{2mm}
\begin{tabular}{llrr}
\multicolumn{2}{c}{Setting}  & Gradient (ms) &  Forward (ms) \\
 \toprule
 \multirow{2}{*}{\rot{CPU}} & Theano (6 thread) & 205.77 & 96.24\\
& Torch (6 threads) & \textbf{117.18} &  \textbf{54.8} \\
 \cmidrule(lr){1-4}
\multirow{2}{*}{\rot{GPU}} & Theano + cuDNN v3 & \textbf{16.72} & \textbf{4.66} \\
& Torch + cuDNN v3 & 98.74 & 29.2\\
\bottomrule
\end{tabular}
\end{table}

\bibliography{DLPaper_KDD2016}
\bibliographystyle{abbrv}


\section{Appendix}\label{sec:appendix}

\subsection{Caffe}\label{ssec:caffe}
Caffe is a deep learning tool developed by the Berkeley Vision and Learning Center and by community contributors and is released under BSD 2-Clause license~\cite{jia14caffe}. It is developed in C\texttt{++} with expression, speed, and modularity in mind which uses CUDA for GPU computation and has commandline, Python, and Matlab interfaces for training and deployment purposes. It separates the definition of the network architecture from actual implementation allowing to conveniently and quickly explore different architectures and layers on either CPU or GPU. Caffe can use LMDB database that allocates memory on the host and device automatically and lazily based on demand for efficient memory usage and high-throughput. The LMDB database supports concurrent reads.

Several types of layers and loss functions are already implemented which can be configured in the form of arbitrary directed acyclic graphs in a configuration file. There are also pre-trained models for popular networks such as “AlexNet” (with non-commercial license) which allows reproducible research. At the time of writing this report, Caffe supports various layers such as convolution, fully connected and pooling layers, etc. The convolution operation can be computed using either a native implementation (by dense matrix multiplications using Blas) or Nvidia cuDNN, if it is installed, where latter usually results in faster computation.  

\subsection{TensorFlow}\label{ssec:tensorflow}
TensorFlow is a C++ based deep learning framework along with python APIs developed and open sourced under an open source Apache 2.0 License by Google recently. TensorFlow uses data flow graphs for performing numerical computations where the nodes represent mathematical operations and the edges represent multidimensional data array communicated between them. TensorFlow has a flexible architecture that supports multiple backends, CPU or GPU on desktop, server or mobile platforms. TensorFlow also offers users the capability to run each node on a different computational device making it highly flexible. Similar to Theano, TensorFlow has automatic differentiation and parameter sharing capabilities which allows a wide range of architectures to be easily defined and executed. TensorFlow has a fast growing community of users and contributors making it an important deep learning framework within the community.

\subsection{Theano}\label{ssec:theano}
Theano is a free Python symbolic manipulation library, under a BSD license, aiming to improve execution time and development time for machine learning algorithms~\cite{bergstra11, bastien12}. It has specifically been utilized for the gradient-based methods such as deep learning that require repeated computation of the tensor-based mathematical expressions. Such mathematical expressions can be rapidly coded in Theano using a high-level description language similar to a functional language that can be compiled and executed on either a CPU or a GPU. 

Theano uses CUDA library to define arrays located on an Nvidia GPU memory with Python bindings. Theano includes many custolllrrm C and CUDA code generators tailored for different types, sizes, and shapes of inputs which optimizes the computation of the complicated tensor computations. 
Theano benefits from a large user community that contribute to its development partly due to the ease of development offered by Python language and its scientific computing stack. Examples of the deep learning algorithms implemented using Theano can be found at~\cite{lisa14}. In the latest version of Theano used here (Theano 0.7), the convolution operation automatically uses the optimized Nvidia cuDNN library, if installed, to perform the convolution. It also provides two additional implementations for the convolution operation, an FFT-based implementation~\cite{mathieu13} and an implementation based on the open-source code of Alex Krizhevsky~\cite{krizhevsky12}.
 While Theano is a general mathematical expression library and may have a relatively steep learning curve for writing efficient code and debugging, several libraries (e.g. Pylearn2, Keras, and Lasagne) have been developed on top it which are specifically tailored for deep learning algorithm providing building blocks for fast experimentation of the well-known methods. 

\begin{table*}[!htb]
\caption{The averaged processing times of the stacked autoencoders (AE) for both pre-training and fine-tuning steps using batch size of 64. The encoder dimensions for AE1, AE2, and AE3 are 800, 1000, and 2000, respectively. For the unsupervised pre-training step, the gradient computation times are reported for the individual AEs along with the total gradient computation. For the supervised fine-tuning step of the stacked enocoders (SE), both gradient computation and forward pass times are reported. Caffe and Neon implementations do not have tied weights.}
\label{tab:sae2}
\centering
\small
\setlength{\tabcolsep}{2mm}
\sisetup{detect-weight=true,detect-inline-weight=math,output-decimal-marker=\textnormal{.}}
\begin{tabular}{lllS[table-format=3.1]S[table-format=3.1]S[table-format=3.1]S[table-format=3.1]S[table-format=3.1]S[table-format=3.1]S[table-format=3.1]}
& &  & \multicolumn{5}{c}{Gradient (ms)} &  {Forward (ms)}\\
 \cmidrule(lr){4-8} \cmidrule(lr){9-9}
\multicolumn{3}{c}{Setting} & {AE1} & {AE2} & {AE3} & {Total pre-training} & {SE} & {SE} \\
\toprule
\multirow{8}{*}{\rot{CPU threads}} & \multirow{3}{*}{\rot{1}} & Neon & 24.6&46.1&120.1&190.8&76.5 &29.3 \\
&& TensorFlow & 44.4 & 65.3 & 126.1 & 235.8 & 131.5 & 69.5 \\
&& Theano & 23.2&36.9 &79.0 &139.1 &65.1 &43.2 \\
&& Torch & 22.9&35.0 &79.2 & 137.1 &61.8 &34.0 \\
 \cmidrule(lr){2-9}
& \multirow{2}{*}{\rot{6}} & Theano & 8.1&13.7 &24.8 &46.6 &24.3 &14.6 \\
&& Torch & \boldentry{3.1}{7.6}&\boldentry{3.1}{13.0} &\boldentry{3.1}{24.9} & \boldentry{3.1}{45.5}& \boldentry{3.1}{22.7}&\boldentry{3.1}{11.4} \\
 \cmidrule(lr){2-9}
 & \multirow{3}{*}{\rot{12}} & Caffe & 17.2&30.0 &63.9 & 111.1&44.3 &32.0 \\
&& TensorFlow & 21.2 & 36.0 & 60.2 & 117.4 & 75.5 & 53.2 \\
&& Theano & 8.9& 15.8 &29.0 & 53.7 &25.9 &15.8 \\
&& Torch & 11.4&19.3 &37.7 & 68.4&31.9 & 16.0\\
\cmidrule(lr){1-9}
 \multirow{4}{*}{\rot{GPU}} & \multicolumn{2}{l}{Caffe + cuDNN v3} &\boldentry{3.1}{0.8} &1.1&\boldentry{3.1}{1.5} &\boldentry{3.1}{3.4} & 1.7& 0.9 \\
& \multicolumn{2}{l}{Neon} &1.1&1.5 & 1.9&4.5 &2.0 &1.0 \\
& \multicolumn{2}{l}{TensorFlow + cuDNN v2} &12.0 & 24.2 & 39.1 & 75.3 & 58.2 & 44.4 \\
& \multicolumn{2}{l}{Theano + cuDNN v3} & 0.9&1.2 &2.2 &4.3& \boldentry{3.1}{1.1}&0.9\\
& \multicolumn{2}{l}{Torch + cuDNN v3} & \boldentry{3.1}{0.8}&\boldentry{3.1}{0.9} &1.8 &3.5&1.5 &\boldentry{3.1}{0.7} \\
\bottomrule
\end{tabular}
\end{table*}

\begin{table*}[!htb]
\caption{The averaged times required on Theano to compile both gradient and forward functions for the studied deep networks. The cuDNN library is used for the GPU measurements. We report two sets of measurements. The first set shows the compilation times when the Theano cache is clear. The second set shows the times required to re-compile the functions. The re-compilation times, which are significantly faster, are more indicative of times required to fine-tune and cross-validate an architecture and thus are more relevant for practical scenarios. We noticed from our experiments that changing hyperparameters (e.g. number of feature maps or convolutional layers) causes only slight changes in the re-compilation times. For more information see: \url{http://deeplearning.net/software/theano/extending/pipeline.html}.}
\label{tab:theanoCompilation}
\begin{threeparttable}
\centering
\small
\setlength{\tabcolsep}{2mm}
\begin{tabular}{llrr}
\multicolumn{2}{c}{Setting}  & First compile (s) & Re-compile (s) \\
 \toprule
 \multirow{3}{*}{\rot{CPU}}
& LeNet                                  & 25.2      &   0.7 \\
& Stacked Autoencoder (small)               & 19.9     &   2.0 \\
& LSTM                                     & 80.1      &  12.7 \\

\cmidrule(lr){1-4}
\multirow{4}{*}{\rot{GPU}}
& LeNet 			                         & 177.7    & 5.0  \\
& AlexNet 			                     & 212.0 & 6.1  \\
& Stacked Autoencoder (small)		             & 106.8 & 2.0  \\
& LSTM                                   & 283.5    & 19.7 \\
\bottomrule
\end{tabular}
\end{threeparttable}
\end{table*}

\subsection{Torch}\label{ssec:torch}
Torch is a scientific computational framework built using Lua that runs on Lua (JIT) compiler~\cite{collobert11torch7}. It has strong CUDA and CPU backends and contains well-developed, mature machine learning and optimization packages. The Tensor libraries that come with it have very efficient CUDA backend and the neural networks (nn) libraries can be used to build arbitrary acyclic computation graphs with automatic differentiation functionalities – i.e. It has a \emph{:forward()} function that computes the output for a given input, flowing the input through the network; and it has a \emph{:backward()} function that will differentiate each parameter in the network w.r.t. the gradient that is passed in. Torch also provides bindings to the latest version of Nvidia cuDNN that gives it access to state-of-art speedups for convolutional operations. The latest version, Torch7, has easy to use multi-GPU support and parallelizing packages that make it very powerful for training deep architectures. Torch has a large community of developers and is being actively used within large organizations like Facebook, Google and Twitter. Specifically, many researchers at NYU and Facebook AI Research (FAIR) lab actively contribute to Torch by making a lot of their code open source. Many companies also have in-house teams to customize Torch for their deep learning platforms that has contributed to its popularity in recent times.

\subsection{Neon}\label{ssec:neon}
Neon is a Python based deep learning framework developed by Nervana. It has recently been open-sourced under an open source Apache 2.0 License. Neon has customized CPU and GPU backends, known as NervanaCPU and NervanaGPU backends, respectively. The NervanaGPU backend consists of kernels written in MaxAs assembler and Python wrappers which is highly optimized for Nvidia’s Maxwell GPUs (e.g. Titan X). The NervanaCPU backend is built on top of python NumPy library. Neon supports commonly used models such as convnets, MLPs, RNNs, and autoencoders. Compared to above three frameworks, Neon is a relatively young framework. Thus, it has not yet been adopted widely within the deep learning community and many of the features already available in the other frameworks, are still under development for Neon. More discussions on the available and missing features of Neon will be provided in the following sections.

\subsection{Supplemental results}\label{ssec:eresults}
The performance of the five frameworks on stacked autoencoder are shown in Table~\ref{tab:sae2} where the encoders of the three AE layers have 800, 1000 and 2000 hidden layers, respectively. The averaged times required on Theano to compile both gradient and forward functions for the studied deep networks are reported in Table~\ref{tab:theanoCompilation}. 

\end{document}